\documentclass{article}
\usepackage{ijcai26}

\usepackage{times}
\usepackage{soul}
\usepackage{url}
\usepackage[hidelinks]{hyperref}
\usepackage[utf8]{inputenc}
\usepackage[small]{caption}
\usepackage{graphicx}
\usepackage{amsmath}
\usepackage{amsthm}
\usepackage{booktabs}
\usepackage{algorithm}
\usepackage{algorithmic}
\usepackage[switch]{lineno}
\usepackage[table]{xcolor}
\usepackage{natbib}
\usepackage{amssymb}
\usepackage{array}
\usepackage{multirow}
\usepackage{makecell}
\usepackage{caption}
\newcolumntype{C}[1]{>{\centering\arraybackslash}p{#1}}

\title{``The Whole Is Greater Than the Sum of Its Parts'': \\ A Compatibility-Aware Multi-Teacher CoT Distillation Framework}

\author{
Jin Cui\thanks{Equal contribution.}$^{1}$\and
Jiaqi Guo\footnotemark[1]$^{2}$\and
Ruixuan Yang$^1$\and
Jiayi Lu$^1$\and
Jiepeng Zhou$^{3}$\and
Jiajun Xu$^4$\and
Jiangcheng Song$^1$\and
Boran Zhao\thanks{Corresponding author.}$^{4}$\and
Pengju Ren$^{1}$
\affiliations
$^1$State Key Laboratory of Human-Machine Hybrid Augmented Intelligence,\\
     Institute of Artificial Intelligence and Robotics, Xi'an Jiaotong University \\
$^2$Nankai University, 
$^3$The Hong Kong University of Science and Technology(Guangzhou)\\
$^4$School of Software Engineering, 
State Key Laboratory of Human-Machine Hybrid Augmented Intelligence,
     Institute of Artificial Intelligence and Robotics, Xi'an Jiaotong University \\
\emails
andycui@stu.xjtu.edu.cn,
\{boranzhao, pengjuren\}@xjtu.edu.cn
}

\begin{document} 

\maketitle

\begin{abstract}

Chain-of-Thought (CoT) reasoning empowers Large Language Models (LLMs) with remarkable capabilities, but typically requires enormous parameter scales. CoT distillation has emerged as a promising paradigm to transfer reasoning prowess into compact student models (SLMs), yet existing approaches often rely on a solitary teacher, capping the student’s potential since individual LLMs often exhibit distinct capability biases and may suffer from catastrophic forgetting. While leveraging diverse teachers seems appealing, effectively fusing their supervisions remains challenging: teacher-student incompatibility risks amplifying hallucinations, and passive supervision fails to ensure genuine logic internalization. To address this, we introduce \textit{COMPACT}, a framework that adaptively fuses supervisions from different teachers by dynamically weighting teacher gradients based on the student’s real-time compatibility evaluated by a multi-dimensional metric: (1) \textit{Graph-based Consensus} to filter misleading rationales by identifying mainstream reasoning paths; (2) \textit{Mutual-Information-based Adaptability} to detect ``epiphany moments'' for genuinely understanding the reasoning process rather than merely imitating; and (3) \textit{Loss-based Difficulty} to assess student receptivity to the teacher's guidance and prevent negative transfer. Our experiments and latent space analysis demonstrate that \textit{COMPACT} effectively integrates diverse reasoning capabilities without damaging the model's original knowledge structure, achieving state-of-the-art performance on various benchmarks while effectively mitigating catastrophic forgetting. Code available: \url{https://github.com/CAG-Research/COMPACT.git}.

\end{abstract}

\section{Introduction}
Large Language Models (LLMs) have demonstrated remarkable capabilities in complex reasoning tasks, particularly when empowered by Chain-of-Thought (CoT) reasoning \citep{wei2022emergent, kojima2022large}. However, this emergent ability typically requires enormous parameter scales, hindering its practical deployment in resource-constrained environments. Consequently, CoT distillation offers a promising solution by transferring the reasoning strategies of LLMs into compact Student Models (SLMs) using teacher rationales as supervision \citep{Ho2022LargeLM,magister2023teaching,hsieh2023distilling}.


However, existing distillation approaches typically constrain SLMs to fit a single teacher-generated rationale. Ignoring reasoning path diversity that often leads SLMs to merely mimic the teacher's superficial style rather than internalizing the underlying logic \citep{Gudibande2023TheFP}, thereby limiting generalization capabilities. While subsequent methods attempt to mitigate this by using multiple rationales to enforce answer consistency \citep{chen2023mcc} or employing decoupled LoRA experts for different reasoning tasks. However, these methods are still limited by their dependence on a single teacher model. This poses a significant limitation: as LLMs often have specializations on specific domains, they inevitably exhibit capability biases and susceptibility to catastrophic forgetting \citep{kirkpatrick2017overcoming, Liu2024MoreTC}. Consequently, no single teacher dominates across all domains. Instead, each model offers unique strengths and specialized knowledge \citep{Jiang2023LLMBlenderEL, Wang2024MixtureofAgentsEL}. This necessitates a multi-teacher distillation framework that cultivates a more robust student.

Despite the appeal of collaborative distillation, effectively synthesizing the reasoning capabilities of different teachers remains challenging. First, a fundamental incompatibility exists between teacher distinctiveness and student adaptability. Recent studies indicate that teachers imprint recognizable ``reasoning signatures'' \citep{Chen2025UnveilingTK, Gudibande2023TheFP}; mismatches in these distributions can severely disrupt the transfer of long-horizon reasoning chains \citep{Wu2025BeyondSL,magister2023teaching}. This issue is exacerbated by inevitable noise in teacher-generated rationales that makes mixing diverse reasoning paths without proper filtering risky. It could amplify hallucinations and degrade the reasoning precision of the distilled models. Second, the definition of an ``optimal teacher'' changes during distillation, depending on the dataset and the student's current learning progress. Existing methods largely remain agnostic to this adaptability. They treat SLMs as passive imitators, lacking mechanisms to ensure genuine logic comprehension rather than superficial pattern mimicry, thus failing to align with the student's evolving receptive capacity. Third, determining the optimal fusion paradigm, specifically between supervision-level integration and parameter-level merging, remains an unresolved dilemma. Supervision-level approaches are intrinsically constrained by the difficulty of quantifying teacher-student compatibility in advance. Conversely, while model merging has been explored as a promising alternative \citep{ilharco2023editing, matena2022merging, mitchell2021fast}, static parameter addition or averaging (e.g., \citep{Shen2025MergeofThoughtD}) often induces conflicts within the parameter space in distillation tasks. This leads to destructive interference or ``cancellation effects'' among diverse supervision signals, resulting in suboptimal performance. These challenges necessitate a more adaptive, student-centric fusion mechanism.

To address these challenges, we propose \textbf{COMP}atibility-\textbf{A}ware Multi-teacher \textbf{C}oT Distilla\textbf{T}ion (\textit{COMPACT}). Unlike static parameter merging that often incurs parameter conflicts, \textit{COMPACT} employs a dynamic weighting mechanism that treats teacher-specific updates as independent task vectors to avoid interferences. This approach allows the student to selectively synthesize gradients from the most compatible and high-quality teachers at the instance level, effectively preventing the ``cancellation effects'' in post-hoc merging. Central to this framework is a \textbf{Multi-Dimensional Evaluation} mechanism designed to quantify teacher-student compatibility through three distinct metrics: (1) Graph-based Consensus, which constructs a semantic similarity graph of teacher outputs to filter isolated noise and identify mainstream reasoning paths; (2) Mutual-Information-based Adaptability, an information-theoretic probe that detects ``epiphany moments'' in the student's learning process to distinguish genuine logic comprehension from rote memorization; and (3) Loss-based Difficulty, which assesses the student's receptivity to a teacher's guidance to prevent negative transfer. Furthermore, to ensure robustness across diverse reasoning paths, we impose a consistency constraint that encourages the student to produce stable predictions regardless of different reasoning patterns.


Extensive evaluations across In-Distribution (ID) and Out-of-Distribution (OOD) benchmarks demonstrate that \textit{COMPACT} achieves state-of-the-art performance, significantly enhancing the multi-dimensional capabilities of distilled SLMs. Furthermore, latent space analysis reveals that our method enables the internalization of teacher knowledge without disrupting the student's original representation space, confirming the model's superior ability to mitigate catastrophic forgetting while absorbing diverse reasoning patterns. Our main contributions are summarized as follows:
\begin{enumerate}
    \item We identify the limitations of single teacher supervision and static parameter merging in CoT distillation, specifically the issues of parameter conflict, and the neglect of students’ evolving adaptability.
    \item We propose an effective multi-teacher distillation framework that leverages student-centroid multi-dimensional evaluation to adaptively internalize teacher capabilities while effectively mitigating catastrophic forgetting.
    \item We creatively introduce Mutual Information (MI) to detect ``epiphany moments'' during distillation, enabling the distinction between genuine logic internalization and superficial pattern mimicry to enhance generalization.
    \item Extensive experiments and latent space analysis confirm that \textit{COMPACT} effectively synthesizes the strengths of diverse teachers, achieving the trade-off between superior performance and stability in preventing forgetting.
\end{enumerate}

\section{Related Work}
\subsection{Chain-of-Thought Distillation from LLMs.}

While Chain-of-Thought (CoT) reasoning has empowered Large Language Models (LLMs) to tackle tasks that previously seem to be impossible, these emergent capabilities are heavily dependent on massive parameter scales. Small Language Models (SLMs) often struggle to actively generate coherent reasoning chains, driving efforts to distill the capabilities of LLMs into compact students \citep{Ho2022LargeLM, magister2023teaching, hsieh2023distilling}. To prevent students from learning superficial shortcuts, methods like SCOTT \citep{wang2023scott} enforce rationale-prediction consistency by contrastive decoding, while \citep{Chen2024LearningTM} proposes maximizing mutual information between prediction and explanation tasks. Beyond single-path imitation, MCC-KD \citep{chen2023mcc} leverages diverse rationales to regularize predictions, and UniCoTT \citep{Zhuang2025UniCoTTAU} proposes a unified framework to transfer knowledge from diverse structural CoTs. Recently, EDIT \citep{Dai2025CaptureTK} utilizes a mistake-driven mechanism to assist students in distinguishing reasoning steps. However, despite these sophisticated mechanisms, existing paradigms predominantly rely on a single teacher model. This reliance confines the student's potential to the boundaries of a specific teacher's knowledge distribution and reasoning biases, necessitating collaborative multi-teacher distillation that leverages complementary expertise.

\begin{figure*}[h]
    \centering
    \includegraphics[width=1\linewidth]{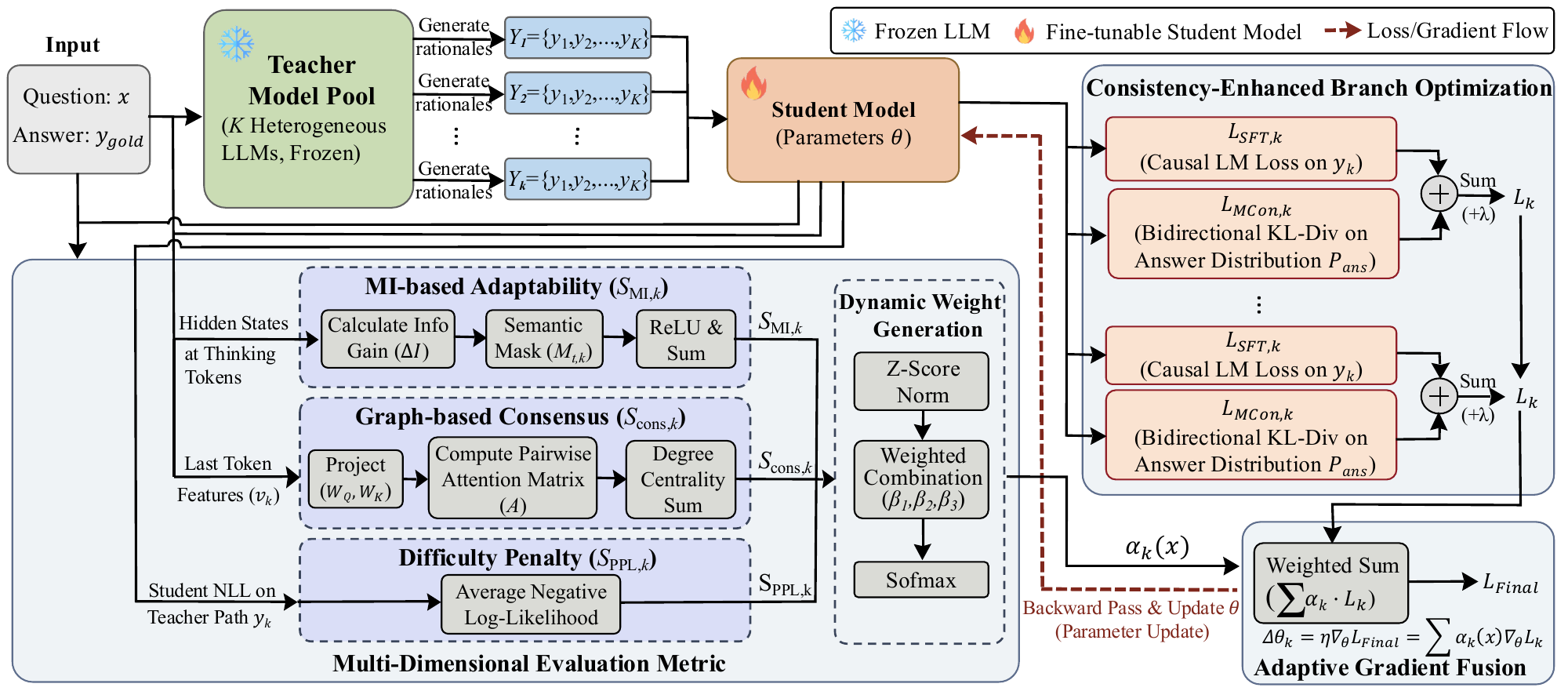}
    \caption{Overview of our \textit{COMPACT} framework. The framework consists of three main components: (1) A frozen Teacher Model Pool generating diverse rationales; (2) A Multi-Dimensional Evaluation Metric that dynamically computes Adaptability ($S_{MI}$), Consensus ($S_{cons}$), and Difficulty ($S_{PPL}$) scores; and (3) An Adaptive Gradient Fusion mechanism that updates the student model using compatibility-aware dynamic weights $\alpha_k(x)$ derived from these scores.}
    \label{fig:main-fig}
    \vspace{-4mm}
\end{figure*}

\subsection{Model Combining and Merging in LLMs.} To break the capability bounds of individual models, recent approaches dynamically combine multiple LLMs during inference. Methods like Mixture-of-Agents \citep{Wang2024MixtureofAgentsEL} fuse intermediate activations, while LLM-Blender \citep{Jiang2023LLMBlenderEL} ranks and ensembles final outputs. However, runtime ensembles inevitably incurs inference overhead and high latency (Time to First Token). Model merging \citep{Yang2024ModelMI} provides a more efficient solution by combining parameters into a single model. Advanced techniques surpass naive averaging by preserving functional capabilities; for instance, Fisher-Weighted Averaging \citep{matena2022merging} utilizes diagonal Fisher information to approximate the posterior distribution, outperforming direct averaging. Similarly, Task Arithmetic \citep{ilharco2023editing} demonstrates that task-specific vectors can be algebraically manipulated to change model behavior, enabling the modular addition or removal of capabilities without retraining. Other works extend these concepts to patch specific model failures \citep{ilharco2022patching, mitchell2021fast} or disentangle weight magnitude from direction to adapt pre-trained models \citep{Yu2024ExtendMM}. However, these merging frameworks usually assume that tasks are orthogonal in parameter space\citep{ilharco2023editing}. This assumption fails in multi-teacher distillation, where different teachers could provide conflicting reasoning paths for the same problem that leads to interference. In contrast, our approach adapts parameter fusion to the distillation setting, aiming to unify diverse reasoning patterns while dynamically resolving interferences between different teacher distributions.

\section{Method}

\subsection{Preliminaries and Problem Formulation}
Let $\mathcal{M}_\theta$ denote the student model parameterized by $\theta$. Given an input question $x$ and a gold answer $y_{gold}$, we assume access to a set of $K$ heterogeneous teacher models. For each input, these teachers generate a set of diverse rationales (CoTs), denoted as $\mathcal{Y} = \{y_1, y_2, \dots, y_K\}$. Our goal is to update $\theta$ by synthesizing the supervision signals from $\mathcal{Y}$. We formulate the parameter update rule based on the concept of Task Arithmetic \citep{ilharco2023editing}. We treat the gradient from each teacher $k$ as a specific ``task vector'' $\Delta \theta_k$. The global update is defined as a dynamic weighted summation:

\vspace{-2mm}
\begin{equation}
\theta_{new} = \theta_{old} + \sum_{k=1}^{K} \alpha_k(x) \cdot \Delta \theta_k
\end{equation}
\vspace{-2mm}

where $\alpha_k(x)$ is a dynamic weighting coefficient conditioned on the student's real-time adaptability to teacher $k$.

\subsection{Multi-Dimensional Dynamic Evaluation}
To determine the optimal weighting coefficients $\alpha_k(x)$, we introduce a multi-dimensional evaluation mechanism, which assesses each teacher's rationale quality from the student's perspective across three dimensions: adaptability, consensus, and difficulty.

\noindent \textbf{MI-based Adaptability Score ($S_{MI}$).} A core challenge in CoT distillation is determining whether a student is genuinely internalizing the reasoning logic or merely mimicking surface textual patterns. Recent findings in information-theoretic analysis of LLMs \citep{Qian2025DemystifyingRD} reveal that the genuine reasoning manifests as ``Information Peaks'' (sudden reductions in the uncertainty of the final answer), occurring specifically at critical ``Thinking Tokens'' (e.g., Therefore, Because, Wait). These tokens act as indicators where the model consolidates prior context to deduce the next logical step.

Based on this, we design $S_{MI}$ to measure the information gain triggered by the teacher's rationale at these specific pivotal moments. We monitor the ``epiphany moments'' where the student's hidden states significantly reduce uncertainty about the final answer $y_{gold}$. For each token $t$ in the reasoning chain $y_k$, we compute the log-likelihood of generating the correct answer using the student's current hidden state $h_{t,k}$:

\vspace{-4mm}
\begin{equation}
I_{proxy}(t,k) = \frac{1}{|y_{gold}|} \sum_{j=1}^{|y_{gold}|} \log P_\theta(y_{gold}^{(j)} | h_{t,k})  
\end{equation}

We define the information gain as $\Delta I_{t,k} = I_{proxy}(t,k)-I_{proxy}(t-1,k)$. To focus on reasoning steps rather than trivial non-reasoning tokens, we apply a semantic mask $M_{t,k}$ which assigns higher weights to predefined ``Thinking Tokens'' (e.g., ``Therefore'', ``So'') and a small weight $\epsilon$ otherwise. We calculate the final adaptability score by aggregating the positive information gains:

\vspace{-2mm}
\begin{equation}
S_{MI,k} = \sum_{t=1}^{T_k} \text{ReLU}(\Delta I_{t,k}) \cdot M_{t,k} 
\end{equation}

Where $T_k$ is the length of the reasoning chain. A high $S_{MI,k}$ indicates that the reasoning path provided by teacher $k$ successfully triggers ``epiphany moments'' in the student's latent space, signaling effective logical transfer.

\noindent \textbf{Graph-based Consensus Score ($S_{cons}$).} Since individual teachers may hallucinate convincing but incorrect reasoning paths, we introduce a consensus score to identify the mainstream logic to filter noises. Unlike external evaluation models, we construct a semantic consensus graph directly within the student's own representation space. This ensures that the selected teacher is not only correct but also aligned with the student's current semantic understanding.

We treat the $K$ reasoning paths as nodes in a fully connected graph. For each path $y_k$, we extract the feature vector $v_k$ from the student's hidden state at the final valid token (EOS). To measure semantic affinity, we project these vectors using the student model's own attention projection matrices ($W_Q$ and $W_K$ from the final transformer layer), which preserves the model's native similarity metric. We compute the pairwise attention matrix $A \in \mathbb{R}^{K \times K}$:

\vspace{-3mm}
\begin{equation}
A_{ij} = \text{Softmax}\left(\frac{(Q_{graph})_i \cdot (K_{graph})_j^T}{\sqrt{d}}\right)  
\end{equation}

Where $Q_{graph} = W_Q v, \quad K_{graph} = W_K v$. The consensus score is derived from the degree Centrality of the graph, defined as the sum of incoming attention weights:

\vspace{-3mm}
\begin{equation}
S_{cons,k} = \sum_{j \neq k} A_{jk}
\end{equation}
\vspace{-3mm}

Paths with high centrality represent the mode of the distribution in the student's semantic space, effectively filtering out outlier hallucinations and provides reliable supervision.

\noindent \textbf{Difficulty Penalty ($S_{PPL}$).} While $S_{MI}$ (although related to student's capability) and $S_{cons}$ evaluate the intrinsic quality of the rationales, they ignore the student's receptivity. A rationale may be logically sound but expressed with a complexity or distribution shift that far exceeds the student's current capacity, making it difficult to absorb effectively. To mitigate this, we introduce a difficulty probe to measure the learnability of each reasoning path. We quantify this using the Negative Log-Likelihood (NLL) of the teacher's rationale $y_k$ evaluated by the student model itself. This serves as a direct proxy for the distributional gap between the teacher's supervision and the student's current knowledge state:

\vspace{-4mm}
\begin{equation}
S_{PPL,k} = -\frac{1}{T_k} \sum_{t=1}^{T_k} \log P_\theta(y_{k,t} | x, y_{k,<t})
\end{equation}
\vspace{-3mm}

A lower $S_{PPL,k}$ indicates the reasoning path lies within the student's ``Zone of Proximal Development'' \citep{Chen2025UnveilingTK}, accessible enough to ensure stable gradient updates while preventing the student from being overwhelmed by excessively perplexing supervision. By integrating this penalty, the framework dynamically lowers the weights of rationales that are incompatible with the student, prioritizing effective knowledge transfer over exposure to complex reasoning.

\noindent \textbf{Dynamic Weight Generation.} The final fusion weight $\alpha_k$ is obtained by normalizing the fusion of these three scores:

\vspace{-2mm}
\begin{equation}
\begin{split}
Score_k = \beta_1 \cdot &\mathcal{N}(S_{MI,k}) + \beta_2 \cdot \mathcal{N}(S_{cons,k}) - \beta_3 \cdot \mathcal{N}(S_{PPL,k}) \\
&\alpha_k = \frac{\exp(Score_k / \tau)}{\sum_{j=1}^K \exp(Score_j / \tau)}
\end{split}
\end{equation}

\noindent where $\mathcal{N}(\cdot)$ denotes Z-Score normalization among teachers.

\subsection{Consistency-Enhanced Branch Optimization}

To prevent the student from overfitting to specific stylistic patterns of any single teacher, we design a dual-objective loss function for each teacher branch $k$.

\noindent \textbf{Supervised Fine-Tuning (SFT).} The primary objective is the standard causal language modeling loss on the teacher's path:

\begin{equation}
L_{SFT,k} = -\sum_{t} \log P(y_{k,t} | x, y_{k,<t})
\end{equation}

\noindent \textbf{Multi-CoT Consistency (MCon).} We enforce a consistency constraint on the answer span. Even if using different reasoning paths, the probability distribution over the final answer $P_{a}$ should remain consistent across high-quality teacher paths. We minimize the Bidirectional KL-Divergence between the current branch $k$ and other branches $j$:

\vspace{-4mm}
\begin{equation}
L_{MCon,k} = \sum_{j \neq k} \alpha_j \cdot D_{KL}\left(P_{a}(\cdot|x, y_k) || (P_{a}(\cdot|x, y_j))\right)
\end{equation}
\vspace{-4mm}

This term encourages the student to achieve outcome-consistency regardless of the reasoning topology. The total loss for branch $k$ is:

\vspace{-2mm}
\begin{equation}
\mathcal{L}_k = L_{SFT,k} + \lambda L_{MCon,k}
\end{equation}

\subsection{Adaptive Parameter Fusion}

Finally, we execute the parameter update. While theoretically formulated as the fusion of task vectors in the parameter space (where $\Delta \theta_k$ is equivalent to $\eta \nabla \mathcal{L}_k$), we implement this efficiently by automatic differentiation on the dynamically weighted objective:

\vspace{-2mm}
\begin{equation}
\mathcal{L}_{Final} = \sum_{k=1}^{K} \alpha_k(x) \cdot \mathcal{L}_k
\end{equation}
\vspace{-2mm}

This generate equivalent gradient aggregation $\nabla_{\theta} \mathcal{L}_{Final} = \sum_{k=1}^{K} \alpha_k(x) \nabla_{\theta} \mathcal{L}_k$ without the overhead of per-branch gradient storage. By optimizing $\mathcal{L}_{Final}$, the student effectively updates its parameters in the direction of interpolated task vectors. This allows the student to prioritize the most instructive and robust reasoning paths for each specific instance.




\section{Experiments}

\begin{table*}[t]
\centering
\scriptsize
\setlength{\tabcolsep}{4.5pt}
\renewcommand{\arraystretch}{1.1}

\definecolor{gain}{RGB}{0,128,128}

\begin{tabular}{l c c c c c c c c}
\toprule
\multicolumn{1}{c}{} &
\multicolumn{4}{c}{\textbf{In-Distribution}} &
\multicolumn{4}{c}{\textbf{Out-Of-Distribution}} \\
\cmidrule(lr){2-5}\cmidrule(lr){6-9}
\textbf{Method} &
\textbf{MATH500} & \textbf{GSM8K} & \textbf{SVAMP} & \textbf{Avg. gain} &
\textbf{CSQA} & \textbf{StrategyQA} & \textbf{GPQA-D} & \textbf{Avg. gain}  \\
\midrule

\rowcolor{gray!15}
\multicolumn{9}{c}{\textbf{Qwen2.5-7B}} \\
\midrule

Base (Qwen2.5-7B-Instruct) & 77.20 & 92.36 & 90.33 & \textcolor{blue}{$\uparrow$4.14} & 83.45 & 68.68 & 30.30 & \textcolor{blue}{$\uparrow$6.69} \\
SBS \citep{hsieh2023distilling} & $77.40_{\textcolor{gain}{\uparrow\,0.20}}$ & $94.77_{\textcolor{gain}{\uparrow\,2.41}}$ & $93.00_{\textcolor{gain}{\uparrow\,2.67}}$ & \textcolor{blue}{$\uparrow$2.38} & $83.20_{\textcolor{red}{\downarrow\,0.25}}$ & $67.25_{\textcolor{red}{\downarrow\,1.43}}$ & $27.46_{\textcolor{red}{\downarrow\,2.84}}$ & \textcolor{blue}{$\uparrow$8.20} \\
MCC \citep{chen2023mcc} & $82.20_{\textcolor{gain}{\uparrow\,5.00}}$ & $90.52_{\textcolor{red}{\downarrow\,1.84}}$ & $91.00_{\textcolor{gain}{\uparrow\,0.67}}$ & \textcolor{blue}{$\uparrow$2.86} & $81.72_{\textcolor{red}{\downarrow\,1.73}}$ & $67.03_{\textcolor{red}{\downarrow\,1.65}}$ & $26.77_{\textcolor{red}{\downarrow\,3.53}}$ & \textcolor{blue}{$\uparrow$9.00} \\
Committee \citep{li2025learningcommitteereasoningdistillation} & $77.25_{\textcolor{gain}{\uparrow\,0.05}}$ & $88.34_{\textcolor{red}{\downarrow\,4.02}}$ & $91.51_{\textcolor{gain}{\uparrow\,1.18}}$ & \textcolor{blue}{$\uparrow$5.07} & $80.52_{\textcolor{red}{\downarrow\,2.93}}$ & $64.19_{\textcolor{red}{\downarrow\,4.49}}$ & $36.6_{\textcolor{gain}{\uparrow\,6.3}}$ & \textcolor{blue}{$\uparrow$7.20} \\
EDIT \citep{Dai2025CaptureTK} & $79.50_{\textcolor{gain}{\uparrow\,2.30}}$ & $94.28_{\textcolor{gain}{\uparrow\,2.49}}$ & $93.50_{\textcolor{gain}{\uparrow\,3.17}}$ & \textcolor{blue}{$\uparrow$1.68} & $83.80_{\textcolor{gain}{\uparrow\,0.35}}$ & $67.50_{\textcolor{red}{\downarrow\,1.18}}$ & $29.10_{\textcolor{red}{\downarrow\,1.20}}$ & \textcolor{blue}{$\uparrow$7.37} \\
MoT\citep{Shen2025MergeofThoughtD} & $81.60_{\textcolor{gain}{\uparrow\,4.40}}$ & $94.37_{\textcolor{gain}{\uparrow\,2.01}}$ & $93.00_{\textcolor{gain}{\uparrow\,2.67}}$ & \textcolor{blue}{$\uparrow$1.14} & $83.20_{\textcolor{red}{\downarrow\,0.25}}$ & $66.81_{\textcolor{red}{\downarrow\,1.87}}$ & $37.88_{\textcolor{gain}{\uparrow\,7.58}}$ & \textcolor{blue}{$\uparrow$5.01} \\
\rowcolor[rgb]{0.886,0.937,0.855}
Ours w/ fusion & $\textbf{82.70}_{\textcolor{gain}{\uparrow\,5.50}}$ & $\textbf{95.30}_{\textcolor{gain}{\uparrow\,2.94}}$ & $\textbf{94.31}_{\textcolor{gain}{\uparrow\,3.98}}$ & -- & $\textbf{83.59}_{\textcolor{gain}{\uparrow\,0.14}}$ & $\textbf{72.93}_{\textcolor{gain}{\uparrow\,4.25}}$ & $\textbf{46.39}_{\textcolor{gain}{\uparrow\,16.09}}$ & -- \\

\midrule

\rowcolor{gray!15}
\multicolumn{9}{c}{\textbf{Llama3.1-8B}} \\
\midrule

Base (Llama3.1-8B-Instruct) & 46.00 & 83.89 & 87.00 & \textcolor{blue}{$\uparrow$4.28} & 74.77 & 70.74 & 21.71 & \textcolor{blue}{$\uparrow$2.68} \\
SBS \citep{hsieh2023distilling} & $52.60_{\textcolor{gain}{\uparrow\,6.60}}$ & $86.96_{\textcolor{gain}{\uparrow\,3.07}}$ & $87.67_{\textcolor{gain}{\uparrow\,0.67}}$ & \textcolor{blue}{$\uparrow$0.84} & $75.02_{\textcolor{gain}{\uparrow\,0.25}}$ & $71.05_{\textcolor{gain}{\uparrow\,0.31}}$ & $19.70_{\textcolor{red}{\downarrow\,2.01}}$ & \textcolor{blue}{$\uparrow$3.16} \\
MCC \citep{chen2023mcc} & $53.00_{\textcolor{gain}{\uparrow\,7.00}}$ & $85.01_{\textcolor{gain}{\uparrow\,1.12}}$ & $83.33_{\textcolor{red}{\downarrow\,3.67}}$ & \textcolor{blue}{$\uparrow$2.80} & $73.33_{\textcolor{red}{\downarrow\,1.44}}$ & $67.99_{\textcolor{red}{\downarrow\,2.75}}$ & $18.53_{\textcolor{red}{\downarrow\,3.18}}$ & \textcolor{blue}{$\uparrow$5.13} \\
Committee \citep{li2025learningcommitteereasoningdistillation} & $51.93_{\textcolor{gain}{\uparrow\,5.93}}$ & $81.83_{\textcolor{red}{\downarrow\,2.06}}$ & $87.84_{\textcolor{gain}{\uparrow\,0.84}}$ & \textcolor{blue}{$\uparrow$2.71} & $74.95_{\textcolor{gain}{\uparrow\,0.18}}$ & $70.87_{\textcolor{gain}{\uparrow\,0.13}}$ & $18.47_{\textcolor{red}{\downarrow\,3.24}}$ & \textcolor{blue}{$\uparrow$3.65}\\
EDIT \citep{Dai2025CaptureTK} & $52.80_{\textcolor{gain}{\uparrow\,6.80}}$ & $86.91_{\textcolor{gain}{\uparrow\,3.02}}$ & $87.67_{\textcolor{gain}{\uparrow\,0.67}}$ & \textcolor{blue}{$\uparrow$0.79} & $74.82_{\textcolor{gain}{\uparrow\,0.05}}$ & $71.15_{\textcolor{gain}{\uparrow\,0.41}}$ & $20.20_{\textcolor{red}{\downarrow\,1.51}}$ & \textcolor{blue}{$\uparrow$3.03} \\
MoT\citep{Shen2025MergeofThoughtD} & $50.80_{\textcolor{gain}{\uparrow\,4.80}}$ & $86.73_{\textcolor{gain}{\uparrow\,2.83}}$ & $85.00_{\textcolor{red}{\downarrow\,2.00}}$ & \textcolor{blue}{$\uparrow$2.40} & $74.94_{\textcolor{gain}{\uparrow\,0.17}}$ & $67.25_{\textcolor{red}{\downarrow\,3.49}}$ & $23.74_{\textcolor{gain}{\uparrow\,2.03}}$ & \textcolor{blue}{$\uparrow$3.11} \\
\rowcolor[rgb]{0.886,0.937,0.855}
Ours w/ fusion & $\textbf{53.56}_{\textcolor{gain}{\uparrow\,7.56}}$ & $\textbf{88.25}_{\textcolor{gain}{\uparrow\,4.36}}$ & $\textbf{87.93}_{\textcolor{gain}{\uparrow\,0.93}}$ & -- & $\textbf{76.26}_{\textcolor{gain}{\uparrow\,1.49}}$ & $\textbf{71.72}_{\textcolor{gain}{\uparrow\,0.98}}$ & $\textbf{27.27}_{\textcolor{gain}{\uparrow\,5.56}}$ & -- \\

\midrule

\rowcolor{gray!15}
\multicolumn{9}{c}{\textbf{Qwen2.5-1.5B}} \\
\midrule

Base (Qwen2.5-1.5B-Instruct) & 50.40 & 72.81 & 79.33 & \textcolor{blue}{$\uparrow$3.25} & 74.20 & 58.07 & 24.75 & \textcolor{blue}{$\uparrow$4.14} \\
SBS \citep{hsieh2023distilling} & $49.80_{\textcolor{red}{\downarrow\,0.60}}$ & $73.91_{\textcolor{gain}{\uparrow\,1.10}}$ & $78.52_{\textcolor{red}{\downarrow\,0.81}}$ & \textcolor{blue}{$\uparrow$3.35} & $73.46_{\textcolor{red}{\downarrow\,0.74}}$ & $56.55_{\textcolor{red}{\downarrow\,1.52}}$ & $22.99_{\textcolor{red}{\downarrow\,1.76}}$ & \textcolor{blue}{$\uparrow$5.48} \\
MCC \citep{chen2023mcc} & $53.60_{\textcolor{gain}{\uparrow\,3.20}}$ & $70.54_{\textcolor{red}{\downarrow\,2.27}}$ & $79.80_{\textcolor{gain}{\uparrow\,0.47}}$ & \textcolor{blue}{$\uparrow$2.78} & $71.32_{\textcolor{red}{\downarrow\,2.88}}$ & $56.99_{\textcolor{red}{\downarrow\,1.08}}$ & $20.71_{\textcolor{red}{\downarrow\,4.04}}$ & \textcolor{blue}{$\uparrow$6.80} \\
Committee \citep{li2025learningcommitteereasoningdistillation} & $58.10_{\textcolor{gain}{\uparrow\,7.70}}$ & $73.69_{\textcolor{gain}{\uparrow\,0.88}}$ & $77.98_{\textcolor{red}{\downarrow\,1.35}}$ & \textcolor{blue}{$\uparrow$0.84} & $71.21_{\textcolor{red}{\downarrow\,2.99}}$ & $59.43_{\textcolor{gain}{\uparrow\,1.36}}$ & $27.07_{\textcolor{gain}{\uparrow\,2.32}}$ & \textcolor{blue}{$\uparrow$3.91} \\
EDIT \citep{Dai2025CaptureTK} & $51.20_{\textcolor{gain}{\uparrow\,0.80}}$ & $74.50_{\textcolor{gain}{\uparrow\,1.69}}$ & $79.60_{\textcolor{gain}{\uparrow\,0.27}}$ & \textcolor{blue}{$\uparrow$2.33} & $74.47_{\textcolor{gain}{\uparrow\,0.27}}$ & $57.22_{\textcolor{red}{\downarrow\,0.85}}$ & $23.74_{\textcolor{red}{\downarrow\,1.01}}$ & \textcolor{blue}{$\uparrow$4.67} \\
MoT\citep{Shen2025MergeofThoughtD} & $54.00_{\textcolor{gain}{\uparrow\,3.60}}$ & $74.45_{\textcolor{gain}{\uparrow\,1.64}}$ & $75.25_{\textcolor{red}{\downarrow\,4.08}}$ & \textcolor{blue}{$\uparrow$2.86} & $71.79_{\textcolor{red}{\downarrow\,2.41}}$ & $58.08_{\textcolor{gain}{\uparrow\,0.01}}$ & $28.71_{\textcolor{gain}{\uparrow\,3.96}}$ & \textcolor{blue}{$\uparrow$3.62} \\
\rowcolor[rgb]{0.886,0.937,0.855}
Ours w/ fusion & $\textbf{56.20}_{\textcolor{gain}{\uparrow\,5.80}}$ & $\textbf{75.08}_{\textcolor{gain}{\uparrow\,2.27}}$ & $\textbf{81.00}_{\textcolor{gain}{\uparrow\,1.67}}$ & -- & $\textbf{75.37}_{\textcolor{gain}{\uparrow\,1.17}}$ & $\textbf{58.73}_{\textcolor{gain}{\uparrow\,0.66}}$ & $\textbf{35.33}_{\textcolor{gain}{\uparrow\,10.58}}$ & -- \\

\bottomrule
\end{tabular}
\vspace{-2mm}
\caption{Performance comparison of COMPACT and other methods. $\textcolor{gain}{\uparrow}$ and $\textcolor{red}{\downarrow}$ indicate the relative improvement/decrease to the base model. Avg. gain calculates the average relative gain of our method toward baselines. All baselines are trained on their original settings.}
\vspace{-4mm}
\label{tab:mian-result}
\end{table*}

\subsection{Experimental Setup}
\textbf{Datasets.}
We evaluate \textit{COMPACT} on several benchmarks to verify performance and generalization: (1) In-Distribution mathematical reasonging benchmarks: MATH500 \citep{lightman2023let}, GSM8K \citep{cobbe2021trainingverifierssolvemath}, SVAMP  \citep{patel2021nlpmodelsreallyable}. (2) Out-Of-Distribution commonsense benchmarks: CSQA \citep{talmor2019commonsenseqa}, StrategyQA \citep{geva2021strategyqa}, GPQA-Diamond \citep{rein2023gpqa}.

\noindent \textbf{Models and Implementation Details.} 
We employ open-source DeepSeek-R1-Llama-70B, Qwen2.5-70B, Llama-3.1-70B and QWQ-32B as our teacher models, selected for their remarkable reasoning performance. For students, we select widely-used Qwen2.5-1.5B, Qwen2.5-7B, and Llama3.1-8B to evaluate scalability and architectural universality. We apply LoRA to all linear layers within the student model's attention and feed-forward modules during distillation. We train the models for 8 epochs(1.5B) and 6 epochs(7B/8B) with a batch size of 4 per device and a learning rate of $1\times 10^{-4}$ for 7B/8B models and $5\times 10^{-5}$ for the 1.5B model. All experiments are conducted on 2 $\times$ NVIDIA A100 (80GB) GPUs. 

\noindent \textbf{Baselines.} We evaluate \textit{COMPACT} against diverse baselines:  (1) Zero-shot base models, specifically Qwen2.5-1.5B-Instruct, Qwen2.5-7B-Instruct, and Llama3.1-8B-Instruct; (2) SbS-KD \citep{hsieh2023distilling}, representing vanilla CoT distillation; (3) MCC-KD \citep{chen2023mcc}, a multi-path distillation approach enforcing consistency; (4) EDIT \citep{Dai2025CaptureTK}, which emphasizes mistake-driven key reasoning steps; (5) MoT \citep{Shen2025MergeofThoughtD}, baseline method that averages parameters over all students fine-tuned on various teachers; (6) Committee \citep{Li2024LearningFC}, a mistakes-aware method that simulates a peer-review among multiple teachers. 


\subsection{Main Results}

\begin{figure*}[h]
    \centering
    \includegraphics[width=1\linewidth]{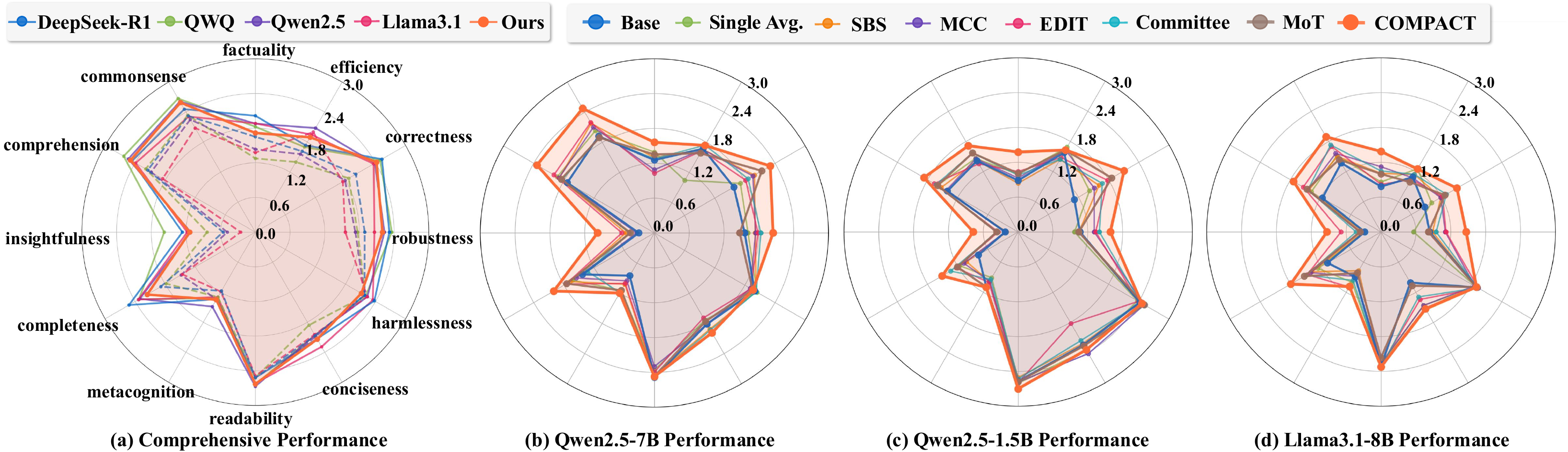}
    \vspace{-5mm}
    \caption{Fine-grained capability evaluation on FLASK framework. (a) compares teachers (DeepSeek-R1-Distill-Llama-70B, QWQ-32B, Qwen2.5-72B, Llama3.1-70B)(solid lines) and their corresponding distilled students (dashed lines), alongside \textit{COMPACT}. (b)(c)(d) show the performance of different distillation methods in Table \ref{tab:mian-result} across three student architectures. All results are scaled for better illustration.}
    \label{fig:flask_eval}
    \vspace{-4mm}
\end{figure*}

\begin{figure}[h]
    \centering
    \includegraphics[width=1\linewidth]{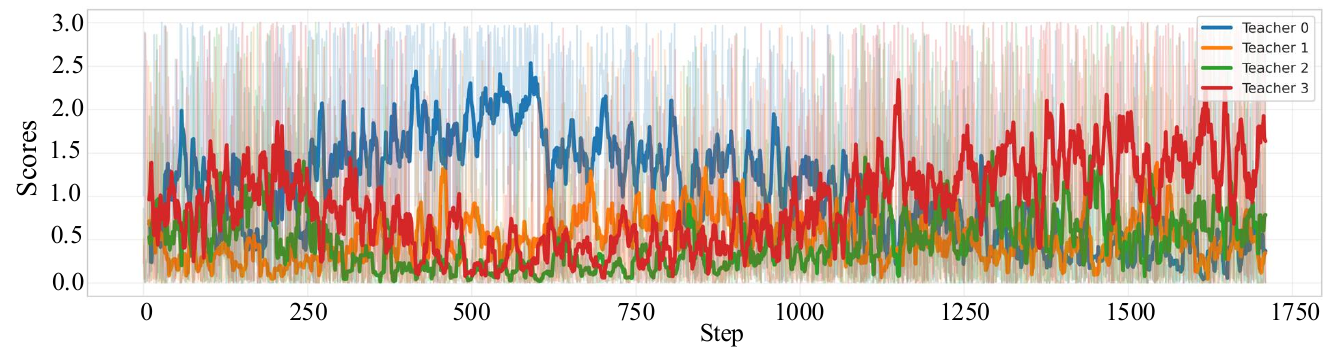}
    \vspace{-6mm}
    \caption{Evolution of dynamic weighting scores for different teachers during training. Teacher 0-3 corresponds to Qwen2.5-72B, Llama3.1-70B, DeepSeek-R1-Llama-70B, and QWQ-32B.}
    \label{fig:score_single}
    \vspace{-4mm}
\end{figure}

Table \ref{tab:mian-result} presents the performance of our method against baselines across three student architectures. \textit{COMPACT} consistently outperforms all baselines on In-Distribution (ID) reasoning tasks. Compared to multi-teacher baselines like MoT and Committee, \textit{COMPACT} demonstrates superior performance and robustness. The performance gains are consistent across diverse model families and scales. Notably, the 1.5B student distilled via \textit{COMPACT} even surpasses the 7B and 8B Base model on GPQA-D, highlighting the method's ability to compress reasoning capabilities into lightweight models efficiently. Crucially, \textit{COMPACT} excels in preserving general capabilities. While baselines often suffer from catastrophic forgetting, our method achieves improvement on evaluated OOD tasks. This confirms that our compatibility-aware weighting effectively isolates reasoning enhancement from general knowledge retention, preventing the tax on versatility often paid by standard distillation. Figure \ref{fig:score_single} shows the dynamic weighting scores for different teachers during training on MATH dataset. Detailed analysis see Appendix C. 

\subsection{Latent Representation Stability Analysis}
To investigate the internal adaptation mechanism of the student model during distillation, we conducted a PCA shift analysis on the latent representations of the layers in Qwen2.5-1.5B distilled by different methods. Following the framework proposed by \citep{Xu2025UnlearningID} and \citep{Wu2025BeyondSL}, we project the high-dimensional activations of the distilled students onto the principal component basis constructed from the original, undistilled student.

\begin{figure}
    \centering
    \includegraphics[width=1\linewidth]{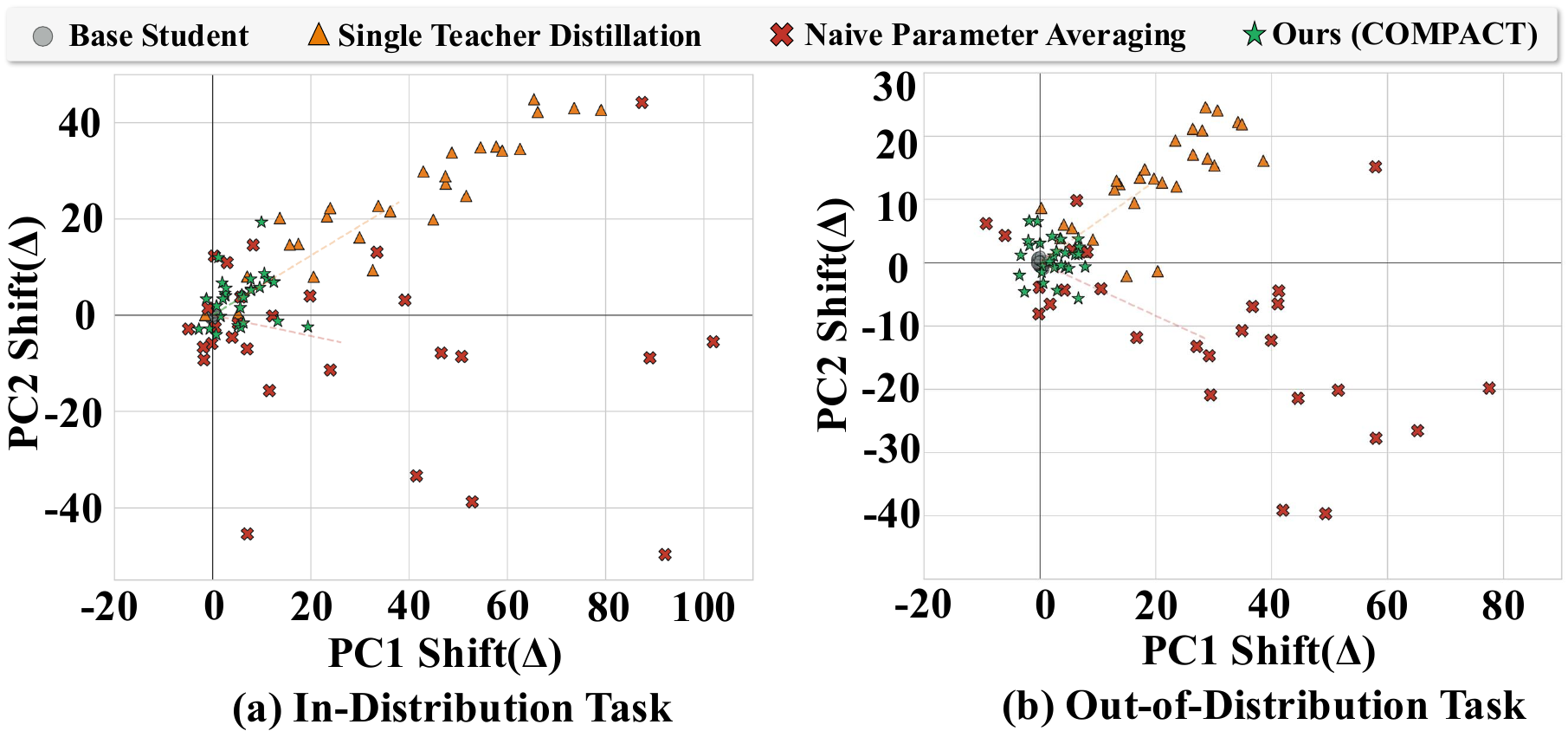}
    \vspace{-6mm}
    \caption{Layer-wise PCA shift of latent representation in Qwen2.5-1.5B distilled by various methods. \textit{COMPACT} exhibits minimal shift, confirming its stability in preserving the general capabilities.}
    \label{fig:pca_shift}
    \vspace{-5mm}
\end{figure}

\noindent \textbf{Methodology and Rationale.} The PCA shift ($\Delta$) measures the euclidean distance between the centroids of latent representations before and after distillation. A large shift, particularly in the first and secondary principal components, indicates a drastic reorganization of the student's feature space, often signaling \textit{structural degradation} or \textit{low corpus affinity}. An ideal distillation should achieve \textit{knowledge absorption without structural destruction}. This requires the student exhibit controlled adaptation on in-distribution tasks (learning reasoning patterns) while maintaining near-zero shift on out-of-distribution tasks (preserving general capabilities).


\noindent \textbf{Analysis on In-Distribution (ID) Reasoning Tasks.} Figure \ref{fig:pca_shift}(a) visualizes the representational shift on the ID reasoning dataset (MATH500). \textit{Naive Parameter Averaging} exhibits the largest and most chaotic dispersion (high shift). This confirms our hypothesis that static parameter averaging induces gradient conflicts among heterogeneous teachers, forcing the student into a ``representational collapse'' where the internal structure is severely twisted to accommodate conflicting reasoning paths. \textit{Single Teacher Distillation} shows a moderate and directional shift, reflecting the ``Style Signature'' effect, where the student is forced to reorganize its cognitive structure to mimic a specific teacher's distribution, potentially at the cost of its original structural stability. In contrast, \textit{COMPACT} maintains a representation centroid tightly aligned with the original student with minimal shift. This suggests that \textit{COMPACT} successfully identifies the most compatible paths, embedding new reasoning capabilities smoothly into the student's existing manifold without destructive reorganization.


\noindent \textbf{Analysis on Out-of-Distribution (OOD) Tasks.} Figure \ref{fig:pca_shift}(b) illustrates the shift on OOD datasets (CSQA), serving as a proxy for catastrophic forgetting. Ideally, the latent representation of OOD tasks should remain invariant to preserve pre-existing knowledge. However, baselines exhibit substantial distributional shifts, indicating their indiscriminate supervision damages the student's original knowledge space. Conversely, \textit{COMPACT} achieves near-zero shift. This confirms that our compatibility-aware weighting successfully disentangles reasoning internalization from general knowledge retention, thereby preserving the student's general capabilities.



\subsection{Fine-grained Capability Analysis}

To conduct a holistic assessment beyond simple accuracy, we employ the FLASK evaluation framework \citep{Ye2023FLASKFL}, which decomposes model capabilities into four dimensions comprising 12 fine-grained skills across four primary abilities: Logical Thinking (\textit{Correctness, Robustness, Efficiency}), Background Knowledge (\textit{Factuality, Commonsense, Metacognition}), Problem Handling (\textit{Comprehension, Completeness, Insightfulness}), and User Alignment (\textit{Readability, Conciseness, Harmlessness}). Results are shown in Figure \ref{fig:flask_eval}.

\noindent \textbf{Breaking Capability Boundaries through Adaptive Fusion.} Figure \ref{fig:flask_eval}(a) confirms that different teachers excel in specific domains and the ``style signature'' phenomenon, where students distilled from individual teachers inherit distinct capability biases.  While the naive parameter averaging (MoT) attempts to combine these abilities and shows gains in \textit{Logical Correctness} and \textit{Completeness}, proving that multi-teacher distillation indeed introduces diverse knowledge, it suffers a critical failure in \textit{Logical Robustness}, accompanied by increased hallucinations (lower \textit{Factuality} and \textit{Commonsense Understanding}). This confirms that conflicting gradient directions neutralize the specialized expertise of individual teachers while disrupting the student's original knowledge structure, ultimately leading to mediocrity. In contrast, leveraging a dynamic weighting mechanism, the \textit{COMPACT}-distilled student effectively breaks these capability boundaries, synthesizing the complementary strengths of diverse teachers to achieve superior performance across all dimensions, even surpassing individual teachers on some metrics.





\noindent \textbf{Robustness and Genuine Comprehension.} Specific capability gains directly validate our multi-dimensional evaluation metrics.
(1) Robustness via Consensus: \textit{COMPACT} holds a significant lead in Factuality and Commonsense Understanding. We attribute this to the Graph-based Consensus ($S_{cons}$) metric, which effectively filters out hallucinations from individual teachers, ensuring the student internalizes only the mainstream correct logic.
(2) Insightfulness via Information Gain: The superior performance in \textit{Insightfulness} and \textit{Completeness} validates the MI-based Adaptability ($S_{MI}$) metric. Unlike baselines that rely on rote memorization of surface patterns, the high mutual information guidance fosters genuine logic comprehension, empowering the student to generate more complete and novel reasoning paths.


\subsection{Ablation Study}

\begin{figure}[h]
    \centering
    \includegraphics[width=1\linewidth]{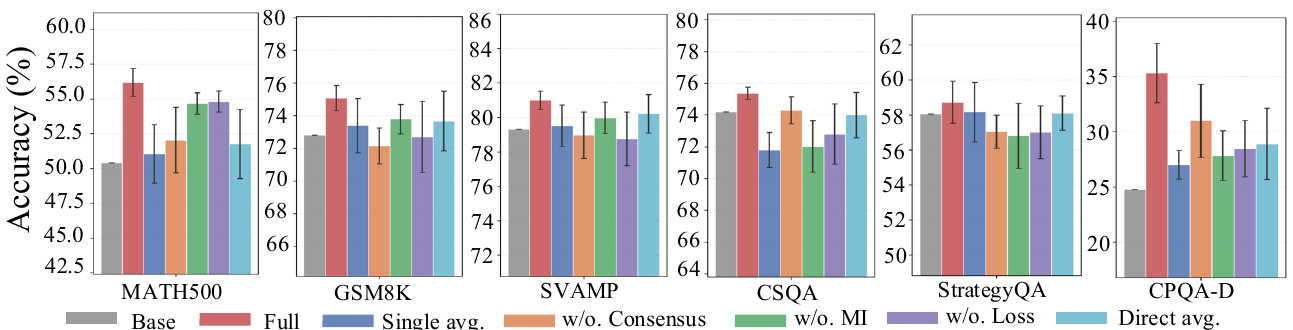}
    \vspace{-6mm}
    \caption{Comprehensive ablation study on model performance.}
    \label{fig:ablation-main}
    \vspace{-7mm}
\end{figure}

To rigorously verify the contribution of each component within our multi-dimensional evaluation metric, we conducted ablation studies on the Qwen2.5-1.5B student model. The results are summarized in Figure \ref{fig:ablation-main}. We calculated the average relative degradation rate (\%) and the shift in stability (Standard Error of the Mean, SEM) across all tasks.

\vspace{-3mm}

\begin{figure}[h]
    \centering
    \includegraphics[width=1\linewidth]{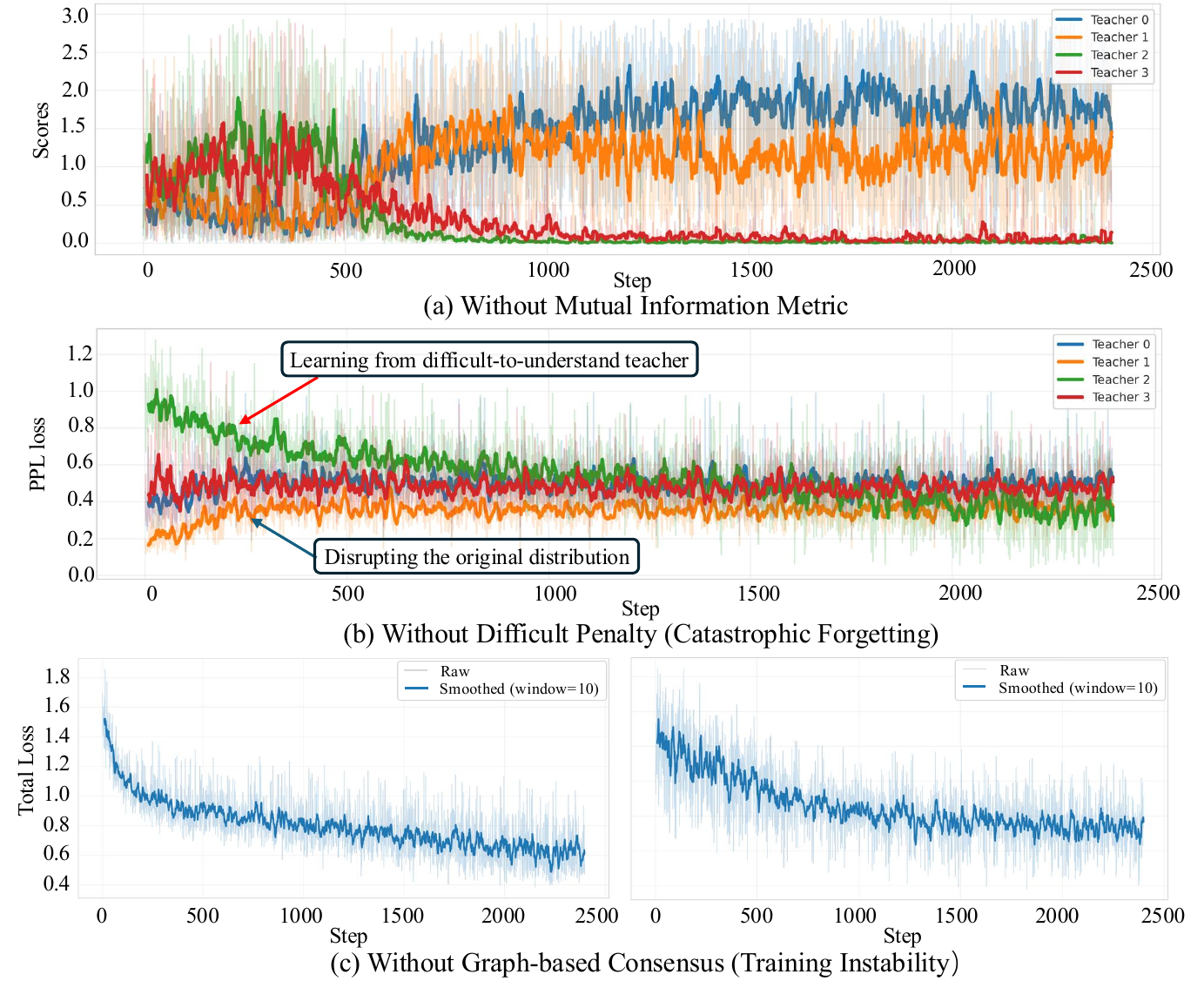}
    \vspace{-6mm}
    \caption{Training dynamics under component ablation. }
    \label{fig:ablation_dynamic}
    \vspace{-3mm}
\end{figure}



\noindent \textbf{Superiority of Dynamic Weighting over Static Merging.} The \textit{``Direct Avg.''} baseline, while improving upon the base model, consistently underperforms the full \textit{COMPACT} framework. Notably, \textit{``Direct Avg.''} baseline exhibits significantly higher variance (e.g., SEM of 1.85 on average). This instability suggests that without compatibility-aware weighting, the student is subject to destructive gradient interference, whereas \textit{COMPACT}’s dynamic weighting stabilizes the learning trajectory by selecting the most compatible paths.

\noindent \textbf{Effectiveness of Graph-based Consensus ($S_{cons}$).} Removing $S_{cons}$ results in a 4.60\% performance drop on ID reasoning tasks and a 5.51\% drop on OOD tasks. More critically, the lack of consensus filtering significantly amplifies the variance of the student's performance: the average SEM doubles ($2.08\times$) on ID tasks and increases by $1.19\times$ on OOD tasks. As illustrated in Figure \ref{fig:ablation_dynamic}(c), the removal introduces significant stochasticity into the optimization landscape that significantly hampers convergence efficiency and leads to an elevated final loss plateau. This validates $S_{cons}$ is essential for mitigating gradient conflicts and ensuring stable convergence.


\vspace{-3mm}
\label{Information Gain Visualization}
\begin{figure}[h]
    \centering
    \includegraphics[width=0.7\linewidth]{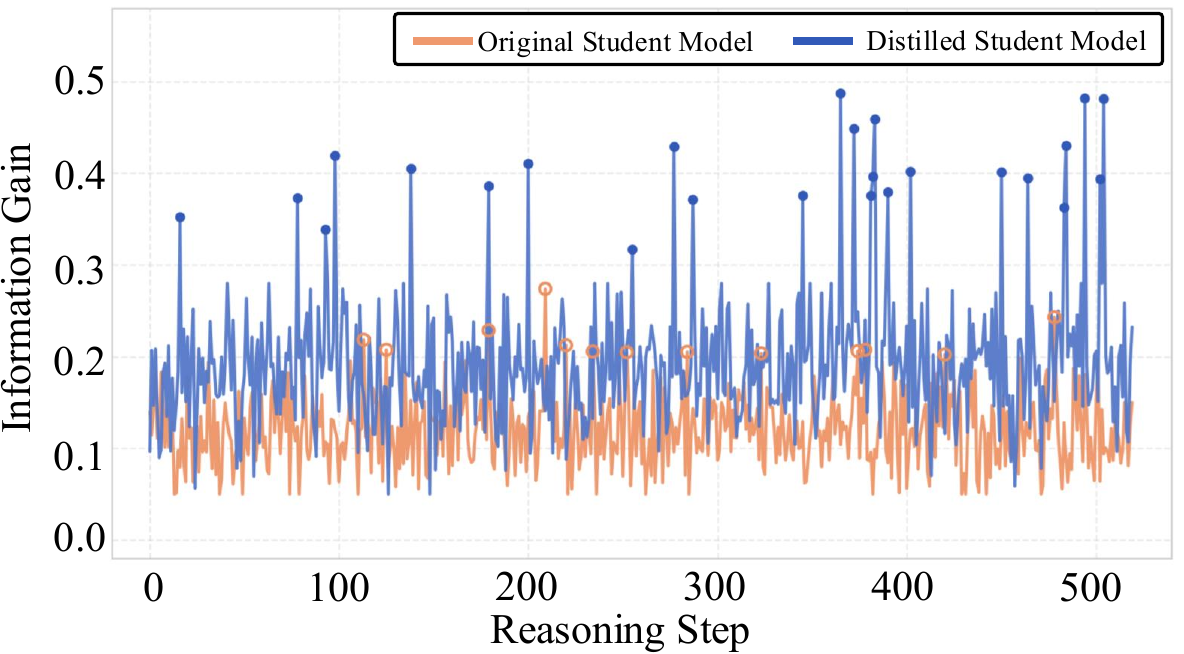}
    \vspace{-3mm}
    \caption{Comparison of mutual information trajectories.}
    \label{fig:MI}
    \vspace{-3mm}
\end{figure}

\noindent \textbf{Effectiveness of MI-based Adaptability ($S_{MI}$).} The removal of the mutual information regularization causes the steepest decline in generalization capabilities, with a significant 9.65\% average drop across OOD tasks and $1.11\times$ SEM increase on ID stability. Figure \ref{fig:ablation_dynamic}(a) also evidences a ``collapse to mediocrity'', where the student model exclusively exploits the low-entropy supervision of less capable generalist teachers (Qwen2.5/Llama3.1) while discarding the high-fidelity but complex reasoning of experts (QWQ/DeepSeek-R1). This confirms that $S_{MI}$ is the decisive factor in prioritizing reasoning depth over mere imitation and the key to the student's insightfulness and generalization capabilities. Figure \ref{fig:MI} confirms our \textit{COMPACT} framework successfully transfers the ``epiphany" capabilities of strong teachers to the student. The distilled student model demonstrates more and sharper peaks while the original student model exhibits a relatively flat and low-entropy trajectory. 



\noindent \textbf{Effectiveness of Loss-based Difficulty ($S_{PPL}$).} Excluding the difficulty penalty ($S_{PPL}$) leads to comprehensive degradation, with an average drop of 2.80\% on ID and 8.60\% on OOD tasks. Similar to consensus, its absence triggers higher volatility: ID variance increases by $1.95\times$ and OOD variance by $1.39\times$. This exclusion also exacerbates catastrophic forgetting (Figure \ref{fig:ablation_dynamic}(b)); the student aggressively overfits to the hard distribution of Teacher 2 at the expense of its proficiency with the generalist teachers, validating the penalty's role as a crucial compatibility adjuster that prevents distributional drift and preserves general capabilities.

\vspace{-1mm}

\section{Conclusion}
In this work, we propose \textit{COMPACT}, a novel compatibility-aware multi-teacher distillation framework that dynamically fuses heterogeneous teacher gradients through instance-level weighting. By integrating multi-dimensional metrics for consensus, adaptability, and difficulty, \textit{COMPACT} successfully synthesizes the complementary strengths of diverse teachers, effectively resolving the ``cancellation effect'' of static merging and minimizing representational drift. Extensive experiments demonstrate that our method achieves state-of-the-art performance with high data efficiency while preserving the student's general capabilities, offering a robust approach for distilling diverse intelligence into compact models.

\section{Acknowledgement}
This work was supported in part by Fundamental and Interdisciplinary Disciplines Breakthrough Plan of the Ministry of Education of China under Grant
JYB2025XDXM504, and National Natural Science Foundation of China No.62302381, No.52441602. The Authors are with the National Key Laboratory of Human-Machine Hybrid Augmented Intelligence and Institute of Artificial Intelligence and Robotics, Xi'an Jiaotong University, Xi'an, Shaanxi, China.


\bibliographystyle{named}
\bibliography{ijcai26}

\appendix

\section{Appendix}
\label{sec:appendix}

\subsection{Details of Tasks and Datasets}

We select MATH500, GSM8K, SVAMP, CommonsenseQA, StrategyQA, and GPQA-Diamond to systematically evaluate model performance across two dimensions: mathematical reasoning and commonsense/knowledge reasoning.
The basic statistics of these benchmarks are presented in
Tables~\ref{tab:math500_subject_stats}--\ref{tab:gpqa_diamond_domain_stats}.

\textbf{MATH500.} MATH500 comprises 500 problems randomly sampled from the MATH dataset~\citep{hendrycks2021measuring}, covering seven subjects across five difficulty levels. Sample distribution in Table \ref{tab:math500_subject_stats} and Table \ref{tab:math500_difficulty_stats}.

\begin{table}[!htbp]
\centering
\small
\renewcommand{\arraystretch}{1.15}
\setlength{\tabcolsep}{10pt}
\begin{tabular}{lcc}
\hline
\textbf{Subject Area} & \textbf{Size} & \textbf{Proportion} \\
\hline
Algebra                  & 124 & 24.8\% \\
Counting \& Probability  &  38 &  7.6\% \\
Geometry                 &  41 &  8.2\% \\
Intermediate Algebra     &  97 & 19.4\% \\
Number Theory            &  62 & 12.4\% \\
Prealgebra               &  82 & 16.4\% \\
Precalculus              &  56 & 11.2\% \\
\hline
\textbf{Total}           & \textbf{500} & \textbf{100.0\%} \\
\hline
\end{tabular}
\caption{Subject-area distribution of MATH500.}
\label{tab:math500_subject_stats}
\vspace{-4mm}
\end{table}

\begin{table}[!htbp]
\centering
\small
\renewcommand{\arraystretch}{1.15}
\setlength{\tabcolsep}{10pt}
\begin{tabular}{lcc}
\hline
\textbf{Difficulty Level} & \textbf{Size} & \textbf{Proportion} \\
\hline
Level 1 &  43 &  8.6\% \\
Level 2 &  90 & 18.0\% \\
Level 3 & 105 & 21.0\% \\
Level 4 & 128 & 25.6\% \\
Level 5 & 134 & 26.8\% \\
\hline
\textbf{Total} & \textbf{500} & \textbf{100.0\%} \\
\hline
\end{tabular}
\caption{Difficulty-level distribution of MATH500.}
\label{tab:math500_difficulty_stats}
\vspace{-4mm}
\end{table}

\textbf{GSM8K.} GSM8K \citep{cobbe2021trainingverifierssolvemath} consists of approximately 8.5K high-quality, linguistically diverse grade school math word problems.

\textbf{SVAMP.} SVAMP \citep{patel2021nlpmodelsreallyable} includes 1,000 one-unknown arithmetic word problems (up to grade 4), constructed by applying structural variations to problem statements. Sample distribution in Table \ref{tab:svamp_test_type_stats}.

\begin{table}[!htbp]
\centering
\small
\renewcommand{\arraystretch}{1.15}
\setlength{\tabcolsep}{10pt}
\begin{tabular}{lcc}
\hline
\textbf{Operation-type} & \textbf{Size} & \textbf{Proportion} \\
\hline
Subtraction       & 160 & 53.33\% \\
Addition          &  59 & 19.67\% \\
Common-Division   &  48 & 16.00\% \\
Multiplication    &  33 & 11.00\% \\
\hline
\textbf{Total}    & \textbf{300} & \textbf{100.00\%} \\
\hline
\end{tabular}
\caption{Operation-type distribution of the SVAMP test set.}
\label{tab:svamp_test_type_stats}
\vspace{-4mm}
\end{table}

\textbf{CommonsenseQA.} CommonsenseQA~\citep{talmor2019commonsenseqa} is a multiple-choice question answering benchmark containing 12,247 examples to test commonsense knowledge. 


\textbf{StrategyQA.} StrategyQA~\citep{geva2021strategyqa} comprises 2,780 questions requiring multi-step strategy inference to answer questions with implicit reasoning steps, making it a challenging benchmark for compositional and knowledge-aware reasoning.

\textbf{GPQA-Diamond.} GPQA-Diamond~\citep{rein2023gpqa} contains 198 expert-written questions in biology, physics, and chemistry, selected for high discrimination between experts and non-experts. Sample distribution in Table \ref{tab:gpqa_diamond_domain_stats}.

\begin{table}[!htbp]
\centering
\small
\renewcommand{\arraystretch}{1.15}
\setlength{\tabcolsep}{10pt}
\begin{tabular}{lcc}
\hline
\textbf{Domain} & \textbf{Size} & \textbf{Proportion} \\
\hline
Chemistry &  93 & 46.97\% \\
Physics   &  86 & 43.43\% \\
Biology   &  19 &  9.60\% \\
\hline
\textbf{Total} & \textbf{198} & \textbf{100.00\%} \\
\hline
\end{tabular}
\caption{Domain distribution of the GPQA-Diamond dataset.  }
\label{tab:gpqa_diamond_domain_stats}
\vspace{-4mm}
\end{table}

\subsection{Multi-perspective Dataset}

We provide a detailed statistical overview of the multi-teacher dataset here. We prompted DeepSeek-R1-Llama-70B, Qwen2.5-70B, Llama-3.1-70B and QWQ-32B to explicitly generate rationales and curated 497 samples from the MATH dataset, each of which contains four valid teacher-generated rationales and a golden answer, strictly following the main text's filtering and stratification criteria. Table ~\ref{tab:math_subject_stats} and Table ~\ref{tab:math_difficulty_stats} summarize the dataset distribution by subjects and difficulty.

\begin{table}[!htbp]
\centering
\small
\renewcommand{\arraystretch}{1.15}
\setlength{\tabcolsep}{10pt}
\begin{tabular}{lcc}
\hline
\textbf{Subject Area} & \textbf{Size} & \textbf{Proportion} \\
\hline
algebra                    & 28 & 14.00\% \\
counting and probability  &  5 &  2.50\% \\
geometry                   & 6   &  3.00\% \\
intermediate algebra       &  12 &  6.00\% \\
number theory              &   33 & 16.50\% \\
prealgebra                 & 102 & 51.00\% \\
precalculus                &   14 &   7.00\% \\
\hline
\textbf{Total}             & \textbf{200} & \textbf{100.0\%} \\
\hline
\end{tabular}
\caption{Subject distribution of the constructed dataset.}
\label{tab:math_subject_stats}
\vspace{-6mm}
\end{table}

\begin{table}[!htbp]
\centering
\small
\renewcommand{\arraystretch}{1.15}
\setlength{\tabcolsep}{10pt}
\begin{tabular}{lcc}
\hline
\textbf{Difficulty Level} & \textbf{Size} & \textbf{Proportion} \\
\hline
Level 1 &  29 & 14.50\% \\
Level 2 &  53 & 26.50\% \\
Level 3 &  45 & 22.50\% \\
Level 4 &  45 & 22.50\% \\
Level 5 &  28 & 14.00\% \\
\hline
\textbf{Total} & \textbf{200} & \textbf{100.0\%} \\
\hline
\end{tabular}
\caption{Difficulty-level distribution of the constructed dataset.}
\label{tab:math_difficulty_stats}
\vspace{-4mm}
\end{table}


\subsection{Experimental Environment}

For brevity, the primary hyperparameter settings and distillation configurations employed in our experiments are listed in Table \ref{tab:hyperparameters}.

\begin{table}[!htbp]
\centering
\small
\begin{tabular}{ll}
\toprule
\textbf{Parameter} & \textbf{Value} \\
\midrule
\textbf{General Settings} & \\
Optimizer & AdamW \\
LoRA Target Modules & All linear layers (Attn. \& FFN) \\
Batch Size (per Device) & 4 \\
Gradient Accumulation & 4 \\
Hardware & 2 $\times$ NVIDIA A100 (80GB) \\
\midrule
\textbf{SLMs Optimization} & \\
Learning Rate (1.5B) & $5 \times 10^{-5}$ \\
Learning Rate (7B/8B) & $1 \times 10^{-4}$ \\
Training Epochs (1.5B) & 8 \\
Training Epochs (7B/8B) &  6 \\
\bottomrule
\end{tabular}
\caption{Settings and hardware implementation.}
\label{tab:hyperparameters}
\vspace{-2mm}
\end{table}

\section{Further Analysis}
\label{sec:Further Analysis}

\begin{figure}[h]
    \centering
    \includegraphics[width=1\linewidth]{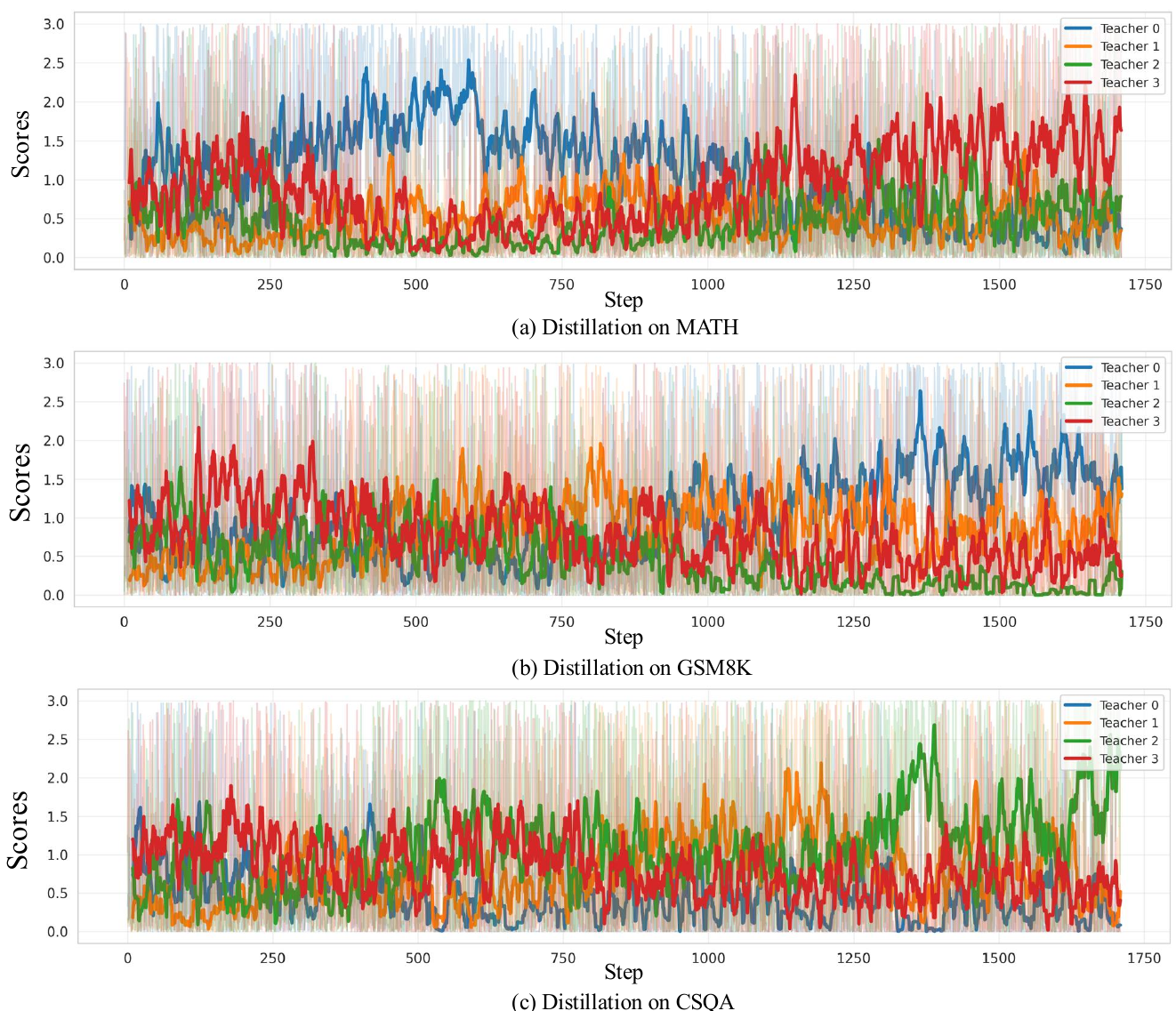}
    \caption{Evolution of dynamic weighting scores for different teachers during training on different datasets. Teacher 0-3 corresponds to Qwen2.5-72B, Llama3.1-70B, DeepSeek-R1-Llama-70B, and QWQ-32B.}
    \label{fig:score}
    \vspace{-2mm}
\end{figure}

\noindent \textbf{Task-Dependent Preference in Mathematical Reasoning.} The training dynamics across mathematical benchmarks elucidate a clear trade-off between reasoning depth and accessibility, governed by task complexity. On the challenging MATH dataset, the student manifests a distinct \textit{curriculum learning} trajectory: it initially prioritizes the less capable generalist teacher (Qwen2.5) to establish foundational logic, before progressively shifting alignment toward the specialized reasoning expert (QWQ) to master intricate reasoning patterns. In contrast, on the elementary GSM8K task, the student eschews the superfluous complexity of stronger reasoners (QWQ/DeepSeek-R1), consistently favoring the simpler and robust solutions of generalist models (Qwen2.5/Llama3.1). This preference confirms that the framework effectively suppresses ``over-reasoning'' in simpler contexts while selectively activating deep reasoning capabilities solely where necessitated by problem difficulty.

\noindent \textbf{Logic Emergence in Commonsense Reasoning.} On the knowledge-intensive CSQA task, the student demonstrates a non-monotonic preference adaptation that eventually absorbs complex reasoning. Unlike the stable dominance observed in math tasks, the student initially oscillates between less capable generalists (Qwen2.5) before converging on the strong reasoning expert (DeepSeek-R1) in the later training stages. This late-stage shift indicates that while generalist models can handle basic knowledge grounding, the superior structural coherence and robust reasoning ability of the strong teacher are more critical for resolving complex implicit reasoning tasks, which the student effectively internalizes only after sufficient training.

\section{Information Gain Visualization}
\label{Information Gain Visualization}
\begin{figure}[h]
    \centering
    \includegraphics[width=0.9\linewidth]{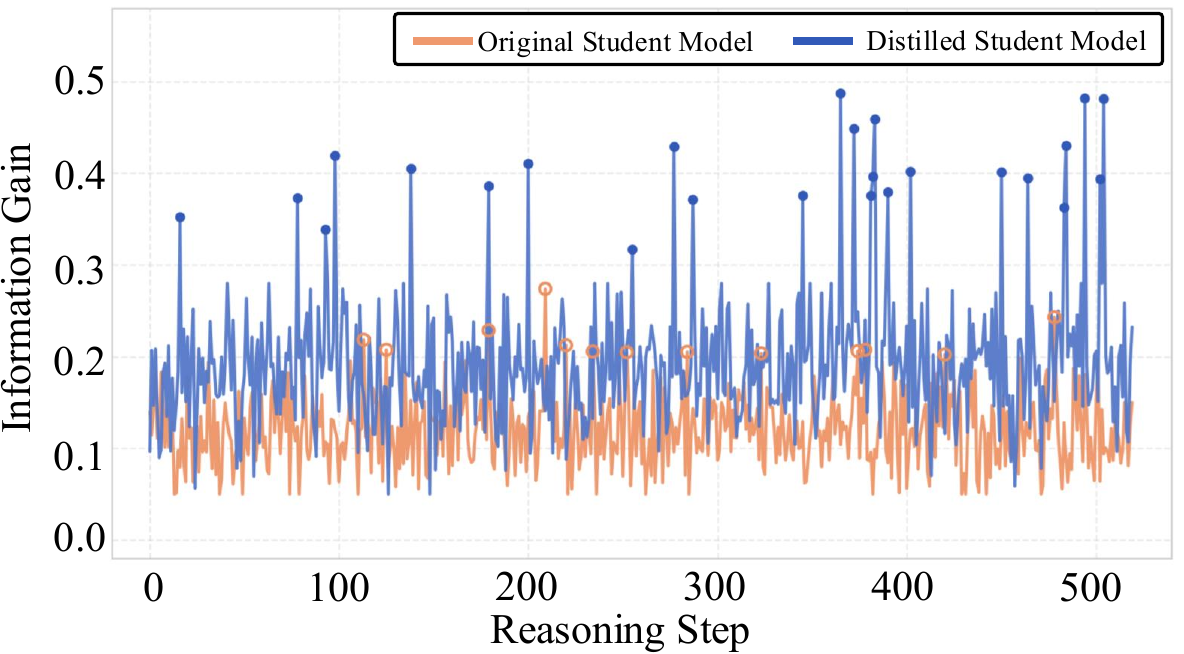}
    \vspace{-2mm}
    \caption{Comparison of mutual information trajectories between the original student model and the \textit{COMPACT}-distilled student model during the reasoning process.}
    \label{fig:MI}
    \vspace{-4mm}
\end{figure}
To verify whether our COMPACT framework successfully transfers the "epiphany" capabilities of strong teachers to the student, we visualize the step-wise Information Gain (MI) on a sample reasoning task. As shown in Figure \ref{fig:MI}, the original student modelexhibits a relatively flat and low-entropy trajectory, suggesting a lack of deep reasoning or logical breakthroughs. In contrast, the distilled student model demonstrates sharp, sparse peaks at specific intervals. These peaks align with the theoretical ``Thinking Tokens'' , confirming that the student has evolved from rote generation to active, structured reasoning with high-information logical transitions.

\section{Distillation Effieciency}
A common concern with multi-teacher distillation is the inference overhead. However, COMPACT effectively offsets this cost through extreme sample efficiency, converging to the SOTA performance in Table \ref{tab:mian-result} using only $~200$ curated samples per teacher (totaling $<1k$ samples). This drastic reduction in dataset size (compared to vanilla methods that typically mandate thousands of examples), completely neutralizes the cost of querying multiple teachers and computing metric scores, resulting in a shorter total training time.

Furthermore, the efficiency advantage is substantial when compared to advanced iterative or reinforcement learning paradigms. Methods like EDIT \citep{Dai2025CaptureTK} and Committee \citep{Li2024LearningFC} require expensive multi-round generation and multi-stage student retraining, while RL approaches like \citep{Patnaik2025LearningTT} necessitate massive rollouts ($>10k$ trajectories). In contrast, COMPACT employs a single-stage dynamic weighting mechanism, eliminating repetitive inference-train loops. This demonstrates that high-quality, compatibility-aware supervision is more computationally efficient than data scaling or complex iterative optimization.

\end{document}